\documentclass[preprint,11pt]{elsarticle}

\usepackage{amssymb}
\usepackage{hyperref}       
\usepackage{url}            
\usepackage{booktabs}       
\usepackage{amsfonts}       
\usepackage{nicefrac}       
\usepackage{microtype}      

\usepackage{epsfig}
\usepackage{graphicx}
\usepackage{color}
\usepackage{amsmath}
\biboptions{sort&compress}
\journal{Neural Networks}

\begin{document}

\begin{frontmatter}



\title{Understanding the Message Passing in Graph Neural Networks via Power Iteration Clustering}


\author[label1]{Xue Li}
\ead{nefu\_education@126.com}

\author[label1]{Yuanzhi Cheng*}
\ead{yzcheng@hitwh.edu.cn}

\address[label1]{ School of Computer Science and Technology\\
  Harbin Institute of Technology\\
  Harbin, Heilongjiang, China 150001 \\}

\begin{abstract}
The mechanism of message passing in graph neural networks (GNNs) is still mysterious. Apart from convolutional neural networks, no theoretical origin for GNNs has been proposed. To our surprise, message passing can be best understood in terms of power iteration. By fully or partly removing activation functions and layer weights of GNNs, we propose subspace power iteration clustering (SPIC) models that iteratively learn with only one aggregator. Experiments show that our models extend GNNs and enhance their capability to process random featured networks. Moreover, we demonstrate the redundancy of some state-of-the-art GNNs in design and define a lower limit for model evaluation by a random aggregator of message passing. Our findings push the boundaries of  the theoretical understanding of neural networks.

\end{abstract}




\end{frontmatter}


\section{Introduction}

The graph neural network (GNN) is one of the most widely used techniques for graph-structured data analysis, with applications in the social sciences, physics, applied chemistry, biology, and linguistics. In virtually every scientific field dealing with graph data, the GNN is the first choice to obtain an impression of one’s data. However, similar to the convolutional neural network (CNN), to explain the mechanism of the GNN is challenging due to its complex nonlinear iterations. It is worth noting that we can understand the GNN better by removing the feature transformations from each layer. This idea can be expressed as
\begin{small}
\begin{equation}
\left\{ 
\begin{array}{l}
X_{1}^{'} = ReLU[MX\Omega_{_{F_{1}}}] \\ 
X_{2}^{'} = ReLU[MX_{1}^{'}\Omega_{_{F_{2}}}] \\ 
                  \quad  \quad \quad ... \\
X_{k-1}^{'} = ReLU[MX_{k-2}^{'}\Omega_{_{F_{k-1}}}] \\
X_{k}^{'} = MX_{k-1}^{'} \\
X^{'} = Softmax(X_{k}^{'} \Omega_{_F})
\end{array}
\right.
\overrightarrow{\ \ yields\ \ } \ 
\left\{ 
\begin{array}{l}
X_{1}^{'} = ReLU[M X] \\ 
X_{2}^{'} = ReLU[M X_{1}^{'}] \\ 
                  \quad  \quad \quad ... \\
X_{k-1}^{'} = ReLU[M X_{k-2}^{'}] \\
X_{k}^{'} = M X_{k-1}^{'} \\
X^{'} = Softmax(X_{k}^{'} \Omega_{_F})
\end{array}
\right.
,
\end{equation}
\end{small}where $M$ is the aggregator, $X$ is the graph feature, and $\Omega$ is the feature transformation matrix.

With the above model simplification, $ReLU$ is invalid due to the fact that the aggregator $M$ is nonnegative and the graph feature can always be transformed to become nonnegative, and every node in a graph will always receive nonnegative information. Then the above k-layer GNN can be expressed as
\begin{small}
\begin{equation}
\left\{ 
\begin{array}{l}
X_{1}^{'} = MX\\ 
X_{2}^{'} = MX_{1}^{'} \\ 
                  \quad  \quad \quad ... \\
X_{k-1}^{'} = MX_{k-2}^{'} \\
X_{k}^{'} = MX_{k-1}^{'} \\
X^{'} = Softmax(X_{k}^{'} \Omega_{_F})
\end{array}
\right.
\overrightarrow{\ \ yields\ \ } \ 
\left\{ 
\begin{array}{l}
X_{k}^{'} = M^{k}X \\ 
X^{'} = Softmax(X_{k}^{'} \Omega_{_F})
\end{array}
\right.
\end{equation}
\end{small}

The expression $X_{k}^{'} = M^{k}X$ is known as the power iteration (without normalization) \cite{ref1}. When $k$ is large enough, multiplying $X$ repeatedly by the matrix $M$ moves every column vector of $X$ to the dominant eigenvector (the eigenvector of the largest-in-magnitude eigenvalue) of $M$. In practice, a GNN is a shallow iteration that calculates an eigenvalue-weighted linear combination of all the eigenvectors of the matrix $M$. For simplicity, we prove the above propositions from a one-dimensional perspective. Assume that the matrix $M$ has $n$ eigenvectors $x_1, x_2, ..., x_n$ with corresponding eigenvalues of $\lambda_1, \lambda_2, ..., \lambda_n$,  in descending order. The $n$ linearly independent eigenvectors form a basis for $R^n$. Then a nonzero random starting vector $v_0$ has
\begin{equation}
	v_0 = c_1 x_1 +c_2 x_2+...+c_n x_n, \quad c_i \ne 0.
\end{equation}
Multiplying both sides of this equation by $M$, we ontain
\begin{equation}
\begin{split}
	M v_0 &= c_1 (Mx_1) +c_2 (Mx_2)+...+c_n (Mx_n)\\
	&= c_1 (\lambda_1 x_1) +c_2 (\lambda_2 x_2)+...+c_n (\lambda_n x_n)
\end{split}
\end{equation}
Repeated multiplication of both sides of this equation by $M$ gives
\begin{equation}
M^k v_0= c_1 (\lambda^k_1 x_1) +c_2 (\lambda^k_2 x_2)+...+c_n (\lambda^k_n x_n)
\end{equation}

The importance of each dimension (eigenvector) is down-weighted by (a power of) its eigenvalue. In spectral clustering, the top $d$ eigenvectors generally define a subspace where the clusters are well-separated. That subspace of  $M^k  v_0$ is somewhat clearer, if we scale the equation by the largest eigenvalue coefficient $c_1 \lambda^k_1 $,
\begin{equation}
\frac{M^k v_0}{c_1 \lambda^k_1 }=  x_1 +\frac{c_2}{c_1} (\frac{\lambda_2}{\lambda_1})^k x_2+...+\frac{c_n}{c_1} (\frac{\lambda_n}{\lambda_1} )^k x_n 
\end{equation}

With increasing $k$, some dimensions shrink quickly and even collapse. Hence we can obtain some ``good" dimensions and diminish the number of ``bad" dimensions. In theory, we can approach the effective subspace by tuning $k$.

The above fact is not new. Lin and Cohen used it to detect communities in an unsupervised way, and proposed power iteration clustering(PIC)  \cite{ref2}, but constrained on one dimension.

When running power iteration with a vector space $M^k [v_0|v_1|...v_p]$, we can capture the true group structure from these varying-convergence trajectories of different vector dimensions. We refer to this as subspace power iteration clustering (SPIC).

Our work makes the following contributions:
\begin{itemize}
	\item [1.]
	 We identify a possible theoretical origin for GNNs apart from CNNs.
	 \item [2.] 
	 We extend GNNs and enhance their capability to process random featured networks by our SPIC models.
	 \item [3.]
	 We classify GNNs and demonstrate the redundancy of current models.
	 \item [4.]
	 We define a lower limit for GNN performance evaluation by random aggregators.
\end{itemize}

The remainder of this paper is organized as follows. Section 2 introduces some related work. SPIC models are proposed in Section 3, and are evaluated and compared in Section 4, where we also discuss experiments to explore their properties. Section 5 provides conclusions and suggestions for future work.

\section{Related Work}
Identifying the static or evolving community structure of networks is drawing increasing attention  \cite{ref3, ref4, ref5}. State-of-the-art GNNs have been tested on the task of community detection. Compared with conventional methods, GNNs show their superior performance at producing nice graph embeddings and capturing complex structures  \cite{ref6} Classic spectral clustering methods such as the normalized cut  \cite{ref7} use the exact eigenvectors to partition the nodes into communities, and they suffer from high computational complexity. GNNs have a similar mechanism with the spectral clustering. The power iteration we explain in the introduction actually performs the matrix decomposition.

When explaining the mechanism of GNNs, people seldom refer to spectral clustering or power iteration clustering. The success of GNNs has been attributed to Laplacian smoothing  \cite{ref8}, which makes the features of vertices in the same cluster similar, and thus easy to cluster. This process can be further understood through power iteration, and is best illustrated in the context of spectral graph drawing.

Here, we take the GraphSAGE-mean model as an example and plot a graph on one dimension, say the $x-axis$. Iteratively placing each node at the average between its old place and the centroid of its neighbors for $k$ times can be expressed as
 \begin{equation}
 (I + D^{-1}A)^{k}x = x',
 \end{equation}
where $D$ is the degree matrix and $A$ is the adjacency matrix.

Combined with the concept of community, by which nodes interact more frequently with members of the same group than with those of other groups, all nodes are thus on the way to their cluster centers. This explains geometrically why power iteration works for community detection. Note that $I+D^{-1}A, D^{-1}A, I - D^{-1}A$ share the same eigenvectors. When $k$ is sufficiently large, $x$ converges to the dominant eigenvector $1_n\equiv (1,1,...,1)^T$ of the degree normalized Laplacian $D^{-1}L$, i.e., all the nodes are put in the same location  \cite{ref9}. 

Some scholars extract a simple Laplacian power model from GNNs but interpret it differently. Wu et al. took the power iteration, $M^kX$, as a feature preprocessing step  \cite{ref10}. Dehmamy et al. regarded $M^k$ as the graph moment, which counts the number of paths from node $i$ to $j$ with length $k$ \cite{ref11}. 

Other interpretations \cite{ref12, ref13, ref14, ref15} focus on graph structures or features, and try to identify the informative components and important node features with a crucial role in a GNN’s prediction. However, when $M$ or $X$ is random, if  $M^k X$ still contributes to the community detection, their interpretations may need some adjustments.

\section{SPIC Models}

We reclassify the Laplacian aggregators, introduce a more general concept, and propose our SPIC models, which are of three types depending on the application of message-Laplacian eigenvalues and eigenvectors. Inspired by the statistical characterization of graph attention networks (GAT) on protein-protein interaction (PPI) data, we enrich these linear models with nonlinear layers.

\subsection{Laplacian Matrix}

Many Laplacians have been proposed, but there is no consensus in the literature as to which definition is most appropriate for message passing. We classify them according to the eigenvectors, since we know the mechanism of GNN:
\begin{equation}
Laplacian \  Aggregator
\left\{ 
\begin{array}{l}
L^{sm} \equiv \{I \pm D^{-\frac{1}{2}}AD^{-\frac{1}{2}}, D^{-\frac{1}{2}}AD^{-\frac{1}{2}} \}\\
L^{rw} \equiv \{I \pm D^{-1}A, D^{-1}A\}\\
L^{di} \ \equiv (\Gamma + \Gamma^T)/2 ,
\end{array}
\right.
,
\end{equation}
where $L^{sm}$ and $L^{rw}$ are symmetric and random-walk Laplacians, respectively. $L^{rw}$ is similar to $L^{sm}$ and $L^{rw}=I-D^{-\frac{1}{2}}(I-L^{sm}D^{-\frac{1}{2}})$. $L^{di}$ is directed Laplacian  \cite{ref16} and $\Gamma$ is an asymmetric weight matrix.

Moreover, we denote $L^G$ as a generalized message Laplacian, by which passing messages contributes to the community detection. 

\subsection{Static Laplacian SPIC}

We propose SPIC models for some state-of-the-art GNNs to show the idea of the static Laplacian. GNNs are listed in Table 1, where $\widetilde{A}=A+I, \widetilde{D}_{ii}= \sum_{j=0}\widetilde{A}_{ij}$, $\Omega^{'}$ is the weight matrix in each GNN layer and $\Omega_{_F}$ is the feature transformation in a linear model.
\begin{table}
 \setlength{\abovecaptionskip}{0pt}
\setlength{\belowcaptionskip}{6pt}
  \caption{Static GNNs}
  \label{table-1}
  \centering
  \begin{tabular}{ll}
    \toprule 
    GNNs     & Aggregator  \\
    \midrule
    GCN  \cite{ref17} & $X' = \widetilde{D}^{-\frac{1}{2}}\widetilde{A} \widetilde{D}^{-\frac{1}{2}}X\Omega^{'}$ \\
    GraphSAGE  \cite{ref18}     & $X' = \widetilde{D}^{-1}\widetilde{A} X\Omega^{'}$  \\
    SGC(linear)  \cite{ref10}     & $X' = (\widetilde{D}^{-\frac{1}{2}}\widetilde{A} \widetilde{D}^{-\frac{1}{2}})^k X\Omega_{_F}$   \\
    \bottomrule
  \end{tabular}
\end{table}
Interestingly, a basic GCN SPIC model, SGC, has been proposed, but its theory is totally different from ours. It misses the power iteration by taking $ (\widetilde{D}^{-\frac{1}{2}}\widetilde{A} \widetilde{D}^{-\frac{1}{2}})^k X$ as a preprocessing step and leaves many questions, which we will answer, such as: why can we remove the activation functions from GNNs, does this removal always work, how to determine $k$, and does only the model redundancy exist? We write this model as
\begin{equation}
DAD(SGC) \equiv (\beta I + \widetilde{D}^{-\frac{1}{2}}\widetilde{A} \widetilde{D}^{-\frac{1}{2}})^k 	X\Omega_{_F} \quad \beta = 0,1,2... 
\end{equation}

GraphSAGE-mean was discussed in Section 2. We directly state it as
\begin{equation}
DA\equiv (\beta I + \widetilde{D}^{-1}\widetilde{A})^k X\Omega_{_F} \quad \beta = 0,1,2... 
\end{equation}

Static Laplacian SPIC is close to the original power iteration and uses one Laplacian matrix as the aggregator.

\subsection{Semistatic Laplacian SPIC}
The aggregators listed in Table 2 suffer from high computational costs and model redundancy. We provide some simple comparisons from experiments, and focus on theoretical analysis.
\begin{table}
\setlength{\abovecaptionskip}{0pt}
\setlength{\belowcaptionskip}{6pt}
  \caption{Semistatic GNNs}
  \label{table-2}
  \centering
  \begin{tabular}{ll}
    \toprule 
    GNNs     & Aggregator  \\
    \midrule
    Spectral GCN  \cite{ref19} & $X' = Ug(\Lambda)U^T X \Omega^{'}$ \\
    ChebNet  \cite{ref20}     & $X' =\sum_{k=0}^{K-1}\Omega_k T_k(\widetilde{L})X$  \\
    TAG  \cite{ref21}    & $X' =\sum_{k=0}^{K}\Omega_k \widetilde{D}^{-\frac{1}{2}}\widetilde{A}^k \widetilde{D}^{-\frac{1}{2}} X\Omega_{_F}$   \\
    APPNP(linear)  \cite{ref22} & $X' = X^{(K)}\Omega_{_F}, X^k = [(1-\alpha)\widetilde{D}^{-\frac{1}{2}}\widetilde{A}^k \widetilde{D}^{-\frac{1}{2}}X^{k-1}+\alpha X_0]$\\
    \bottomrule
  \end{tabular}
\end{table}

The SPIC model of spectral GCN provides a new understanding of its convolutional operations. Removing all the activation functions and the layer weights, we have
\begin{equation}
X' = U g(\Lambda_0) ... g(\Lambda_k) U^T X \Omega_{_F} \approx U	g(\Lambda)^k U^T X\Omega_{_F}
,
\end{equation}
where $U$ is an eigenvector matrix and $g(\Lambda)$ denotes a diagonal eigenvalue matrix.

The eigenvectors are invariant, and the eigenvalues are dynamic. In essence, it calculates a learned eigenvalue-weighted linear combination of the eigenvectors at a high computational expense.

The SPIC model of ChebNet and TAG-like algorithms has been proposed as APPNP. It is derived from PageRank, which is a kind of power method. The creators of APPNP do realize this, but take it as a tool, like SGC. Let us expand them with $K=3$ and $\alpha \in (0,1)$:
\begin{small}
\begin{equation}
\left\{ 
\begin{array}{l}
X_{Cheb}^{'} = [\Omega_0 I + \Omega_1 L_{sm}+\Omega_2 L^{2}_{sm}]X \\
X_{TAG}{'} = [\Omega_0 D^{-1} + \Omega_1 L_{sm}+\Omega_2 L^{2}_{sm}+\Omega_3 L^{3}]X \\
X_{APPNP}^{'} = [\alpha+\alpha(1-\alpha)L_{sm}+\alpha(1-\alpha)^2 L_{sm}^2+(1-\alpha)^3 L_{sm}^3]X\Omega_{_F}
\end{array}
\right.
\end{equation}
\end{small}

They use a similar aggregator, which is a linear combination of DAD(SGC) models. If they do not outperform corresponding static models(GCN/DAD), then we may say the semistatic model has redundancy.

One-dimension APPNP can clarify the definition of the semistatic Laplacian. We denote the starting vector $v_0$ in eigenspace as $v_0 = c_1^{(0)}x_1 + c_2^{(0)}x_2+...+c_n^{(0)}x_n, c_i \ne 0$, and write an APPNP of three iterations as
\begin{small}
\begin{equation}
\begin{split}
X_{APPNP}^{'} &= \sum_{i=0}^{K} (c_1^{(i)}\lambda_{1}^{i} x_1 + c_2^{(i)}\lambda_{2}^{i} x_2+...+c_n^{(i)}\lambda_{n}^{i} x_n)\\
& = g(\lambda_{1})x_1 + g(\lambda_{2})x_2 + ... +g(\lambda_{n})x_n
\end{split}
\end{equation}
\end{small}

The eigenvectors are invariant, and the scaling factor $g(\lambda)$ is a mixture of eigenvalues, which is similar to the SPIC model of Spectral GCN. That is the key of the semistatic models.

\subsection{Dynamic Laplacian SPIC}

The above models are based on the traditional Laplacian, which use no prior information. We present some prior Laplacian models in this section.
\begin{table}
\setlength{\abovecaptionskip}{0pt}
\setlength{\belowcaptionskip}{6pt}
  \caption{Dynamic GNNs}
  \label{table-3}
  \centering
  \begin{tabular}{ll}
    \toprule 
    \quad  GNNs \quad\quad     & \quad \quad Aggregator  \\
    \midrule
    AGNN  \cite{ref23} \quad\quad & \quad\quad $X' = P'X$ \quad \\
    GAT  \cite{ref24} \quad\quad     & \quad\quad $X' = Q'X$ \quad  \\
    \bottomrule
  \end{tabular}
\end{table}
The models in Table 3 use the attention mechanism and integrate the learned feature similarity into their edge weight. The aggregators are dynamic and graph-dependent.

AGNN defines its aggregator as $P_{ij}^{'} = \frac{exp(\varepsilon^{'} \cdot cos(x_i, x_j))}{\sum_{t \in N(i) \cup \{i\}} exp(\varepsilon^{'} \cdot cos(x_i, x_t))}$. $P'$  is symmetric, and there is only one parameter $\varepsilon^{'}$ in each layer. We design its SPIC model by removing all activation functions and feature preprocessing operations to obtain
\begin{equation}
P\_AGNN \equiv (\beta I + P)^k X \Omega_{_F}, \quad \beta=0,1,2...	,
\end{equation}
where $P_{ij}=softmax(\varepsilon\cdot cos(x_i, x_j))$, and $\varepsilon$ is a hyperparameter set to 1.0 in this paper.

GAT is an interesting method. Its attention mechanism causes the relative importance of nodes to differ, which transforms the undirected graph to a bidirectional network with asymmetric edge weights. An asymmetric matrix may not satisfy the diagonalizable condition of the original power iteration. We can symmetrize the attention weight by averaging the matrix and its transpose. That is the idea of the directed Laplacian $L^{di}$.  We create a SPIC model by removing all activations, layer weights, and multi-heads, and iteratively learning with only one attention,
 \begin{equation}
P\_GAT \equiv (\beta I + Q)^k X \Omega_{_F}, \quad \beta=0,1,2...	,
\end{equation}
where $Q = \frac{Z+Z^T}{2}, Z_{ij} = \frac{exp(LeakyReLU(a^T[\Omega_{_F} x_i || \Omega_{_F}x_j]))}{\sum_{t \in N(i) \cup \{i\}} exp(LeakyReLU(a^T[\Omega_{_F} x_i || \Omega_{_F}x_t]))}$, and $a$ is the attention vector.
We design an asymmetric model by directly using the attention weight,
 \begin{equation}
 P\_GAT\_am \equiv (\beta I + Z)^k X \Omega_{_F}, \quad \beta=0,1,2...
 \end{equation}
 
We will further discuss the symmetric or diagonalizable issue below.

\subsection{SPIC with Nonlinear Layers}

We can use GAT to infer the graph types based on the attentions learned. Li et al.  \cite{ref25} observed that the attention weights almost distribute uniformly on all the benchmark citation networks, regardless of the heads and layers. Significant differences are observed for the case of PPI. We classify the above data as linear or nonlinear. For nonlinear data, we need to add some nonlinear layers to our SPIC models. Three testing models based on P\_GAT are as follows,
\begin{equation}
P\_GAT\_Relu1 \equiv
\left\{ 
\begin{array}{l}
X = X\Omega_{_p}\\
X = ReLU(QX) + \beta X \\
X' = (\beta I + Q)^{k-1} X\Omega_{_F}
\end{array}
\right.
\end{equation}
\begin{equation}
P\_GAT\_General \equiv
\left\{ 
\begin{array}{l}
X^{0} = X\Omega_{_p}\\
X^{k} = ReLU(QX^{k-1}\Omega_{_R})+\beta X^{k-1}\\
X' = X^k \Omega_{_F}
\end{array}
\right.
\end{equation}
\begin{equation}
P\_GAT\_w \equiv
\left\{ 
\begin{array}{l}
X = X\Omega_{_p} \\
X' = (\beta I +Q)^k X \Omega_{R}^{k} \Omega_{_F}
\end{array}
\right.
\end{equation}

$ReLU$ is put on the first layer of P\_GAT\_Relu1 to keep the features nonnegative. $Q$ and $X$ are all nonnegative. We put $ReLU$ and another feature transformation $\Omega_{_R}$ in each layer of P\_GAT\_General to strengthen its learning ability. P\_GAT\_w is designed as the linear model of P\_GAT\_General.

By setting $Q = \widetilde{D}^{-\frac{1}{2}}\widetilde{A} \widetilde{D}^{-\frac{1}{2}}$, we can propose DAD\_Relu1, DAD\_General and DAD\_w. We next explore the nonlinear issue by testing these models on PPI.

\section{ Experiment and Exploration}

We compare SPIC models and GNNs on citation networks, conduct experiments to explore the properties of SPIC, and answer the questions posed in Section 3.

\subsection{Datasets and Codes}
We focus on the task of node classification by using citation networks  \cite{ref26} and PPI  \cite{ref27} data. All the citation networks are PyTorch built-in data, which are split well for training. For PPI, we choose two of its 24 networks and treat them as one big network. Dataset statistics are summarized in Table 4.

All experiments on GNNs are performed based on the codes released by PyTorch. Our models and training settings may be found at {\color{blue}https: //github.com/Eigenworld/SPIC }. Results are averaged over 10(for semistatic) or 20 runs, and 100 epochs per run. For the single-label task (e.g., tests on citation networks), we report the mean classification accuracy (with standard deviation), and for the multi-label task (e.g., tests on PPI), we report the micro-averaged F1 score.

\begin{table}
\setlength{\abovecaptionskip}{0pt}
\setlength{\belowcaptionskip}{6pt}
  \caption{Dataset statistics of the citation networks and PPI}
  \label{table-4}
  \centering
  \begin{tabular}{llllll}
    \toprule 
    Type     & Dataset &\#Nodes &\#Edges &Train/Val/Test  &Connected \\
    \midrule
    Linear & Cora & 2,708 & 5,429 &140/500/1,000 &\quad No\\
    Linear & CiteSeer &3,327 & 4,732 &120/500/1,000 &\quad No\\
    Linear & PubMed &19,717 & 44,338 &60/500/1,000 &\quad Yes\\
    Nonlinear & PPI &2,599 & 27,189 &2,050/297/252 &\quad No\\
    \bottomrule
  \end{tabular}
\end{table}

\subsection{Comparison of GNNs and SPIC models}

Many structures of the existing GNNs such as activation functions, layer weights, and multi-aggregators may not be the must-have modules. For fairness, all GNNs compute 64 hidden features and all linear models are iterated for two or three times. Table 5 shows that removing the activations and layer weights from GNNs does not degrade the performance on citation networks. In fact, linear models perform comparably (boldface) to state-of-the-art GNNs. The results of two linear GAT show no big difference between symmetric and asymmetric aggregators. In Section 4.4, we will further test this issue. TAG and APPNP do not show superior performance over GCN and DAD, respectively, which verifies their model redundancy, as mentioned in Section 3.3.

\begin{table}
\setlength{\abovecaptionskip}{0pt}
\setlength{\belowcaptionskip}{6pt}
  \caption{Test accuracy (\%) on citation networks}
  \label{table-5}
  \centering
  \resizebox{\textwidth}{!}{
  \begin{tabular}{llllllll}
    \toprule
    \multicolumn{2}{c}{Model}  &\multicolumn{2}{c}{Cora}&\multicolumn{2}{c}{CiteSeer}&\multicolumn{2}{c}{PubMed}       \\
    \midrule
   GCN & \bf{DAD(SGC)} & $82.2\pm0.6\%$ & $\bf{82.3\pm0.5\%}$  & $72.0\pm1.1\%$ & $\bf{72.0\pm0.4\%}$ &$78.8\pm0.5\%$ & $\bf{79.2\pm0.4\%}$ \\
   SAGE & \bf{DA} & $82.3\pm0.9\%$ & $\bf{82.3\pm0.5\%}$  & $71.4\pm1.0\%$ & $\bf{72.3\pm0.2\%}$ &$78.5\pm0.5\%$ & $\bf{79.3\pm0.7\%}$ \\
    AGNN & \bf{P\_AGNN} & $81.5\pm0.7\%$ & $\bf{82.5\pm0.6\%}$  & $71.5\pm0.7\%$ & $\bf{72.5\pm0.5\%}$ &$78.9\pm0.7\%$ & $\bf{79.0\pm0.6\%}$ \\
     GAT& \bf{P\_GAT} & $82.4\pm0.8\%$ & $\bf{81.7\pm0.5\%}$  & $71.7\pm0.8\%$ & $\bf{71.2\pm1.2\%}$ &$78.1\pm0.6\%$ & $\bf{77.3\pm1.1\%}$ \\
     \multicolumn{2}{c}{\bf{P\_GAT\_am}}&\multicolumn{2}{c}{$\bf{81.0\pm0.7\%}$}&\multicolumn{2}{c}{$\bf{71.0\pm0.8\%}$}&\multicolumn{2}{c}{$\bf{77.3\pm1.0\%}$}\\
     TAG& \bf{APPNP} & $82.4\pm0.9\%$ & $\bf{82.4\pm0.7\%}$  & $71.2\pm0.9\%$ & $\bf{72.0\pm0.3\%}$ &$78.7\pm0.4\%$ & $\bf{78.8\pm0.7\%}$ \\
    \bottomrule
  \end{tabular}}
\end{table}
Moreover, all GAT layers and heads learn similar aggregators on citations networks  \cite{ref25}. These duplicated aggregators are simply shifted versions of each other, and they indicate the linearity of model and data. This may explain why all the activations and layer weights can be removed from GNNs on citation networks. 
\begin{table}
\setlength{\abovecaptionskip}{0pt}
\setlength{\belowcaptionskip}{6pt}
  \caption{Test Micro F1 Score (\%) on PPI}
  \label{table-6}
  \centering
   \resizebox{\textwidth}{!}{
  \begin{tabular}{llllll}
    \toprule 
    Model     & \bf{GAT} &P\_GAT &P\_GAT\_Relu1 &\bf{P\_GAT\_General}  &P\_GAT\_w \\
    PPI &  $\bf{65.5\pm0.5\%}$  &  $51.0\pm1.0\%\downarrow$  &  $54.8\pm1.8\%$  & $\bf{63.6\pm0.5\%}$ & $53.2\pm0.8\%$ \\
    \midrule
    Model  & \bf{GCN} &DAD(SGC) &\bf{DAD\_Relu1} &\bf{DAD\_General}  &DAD\_w \\
    PPI&  $\bf{62.1\pm0.6\%}$  &  $46.1\pm1.0\%\downarrow$  &  $\bf{64.4\pm0.5\%}$  & $\bf{64.1\pm0.7\%}$ & $54.1\pm1.0\%$ \\
    \bottomrule
  \end{tabular}}
\end{table}

Before diving into the tests on nonlinear data, let us observe the attention weights of GAT in Fig. 1, where about eight different attentions are learned. The first two layers capture the similar attention pattern and four different attentions are learned. The attentions captured by the final layer are obviously different from those of previous layers. Pure linear models shown in Table 6 do not work this time. When adding $ReLU$ to the first iteration, DAD\_Relu1 behaves closely to GAT. When adding the activation and layer weight to each iteration, P\_GAT\_General almost reverts to the performance of GAT. The final contrast model shows that the nonlinear layers work, and layer weights contribute slightly.
\begin{figure}
  \centering
  \includegraphics[width=1.0\linewidth]{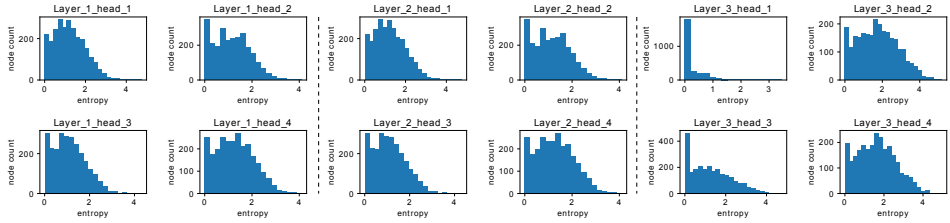}
  \caption{Entropy histogram plots for attention weights of GAT on PPI. Each node entropy is calculated by $H(\{w_{ij}\| j\in N(i)\}) = -\sum_{j\in N(i)} w_{ij} log w_{ij}$.}
\end{figure}

\subsection{Graph Feature Space Exploration}

How seriously does our project suffer from feature redundancy? We show feature redundancy using Cora’s first 800 dimensional features and 193 compressed dimensional features. The original feature size is 1433.

\begin{table}
\setlength{\abovecaptionskip}{0pt}
\setlength{\belowcaptionskip}{6pt}
  \caption{Test accuracy (\%) on reduced Cora}
  \label{table-7}
  \centering
  \resizebox{\textwidth}{!}{
  \begin{tabular}{llllll}
    \toprule
    \multicolumn{2}{c}{Model}  &\multicolumn{2}{c}{Cora\_800}&\multicolumn{2}{c}{Cora\_193}     \\
    \midrule
   GCN & \bf{DAD(SGC)} & $80.6\pm0.5\%$ & $\bf{82.0\pm0.4\%}$  & $75.5\pm1.2\%$ & $\bf{79.0\pm0.6\%}$\\
   SAGE & \bf{DA} & $80.9\pm0.5\%$ & $\bf{81.6\pm0.7\%}$  & $76.2\pm0.9\%$ & $\bf{79.2\pm0.4\%}$ \\
    AGNN & \bf{P\_AGNN} & $80.7\pm0.7\%$ & $\bf{81.5\pm0.3\%}$  & $78.5\pm0.8\%$ & $\bf{79.1\pm0.5\%}$ \\
     GAT& \bf{P\_GAT} & $80.6\pm0.5\%$ & $\bf{80.4\pm0.4\%}$  & $77.4\pm0.6\%$ & $\bf{78.6\pm0.8\%}$  \\
    \bottomrule
  \end{tabular}}
\end{table}
Table 7 shows that we loss little by reducing feature dimensions, especially for SPIC models, which means we can further optimize the graph features. A follow-up question is how can we design the feature width. We explore this issue by running DAD on random-feature graphs. 
\begin{table}
\setlength{\abovecaptionskip}{0pt}
\setlength{\belowcaptionskip}{6pt}
  \caption{Test accuracy (\%) of DAD on random citation networks}
  \label{table-8}
  \centering
  \resizebox{\textwidth}{!}{
  \begin{tabular}{lllll}
    \toprule 
    \bf{Cora\_100} &\bf{Cora\_300} &Cora\_500 &Cora\_1000  &Cora\_2000\\
    $\bf{68.5\pm0.6\% \uparrow}$  &  $\bf{74.5\pm0.7\% \uparrow}$  &  $73.6\pm0.3\%$  & $72.8\pm0.5\%$ & $72.4\pm0.5\%$ \\
    \toprule 
    \bf{Cite\_100} &\bf{Cite\_300} &\bf{Cite\_500} &\bf{Cite\_1000}  &Cite\_2000\\
    $\bf{42.4\pm1.0\%\uparrow}$  &  $\bf{47.0\pm0.7\%\uparrow}$  &  $\bf{48.8\pm0.9\%\uparrow}$  & $\bf{50.9\pm0.6\%\uparrow}$ & $49.6\pm0.9\%$ \\
    \toprule 
    \bf{Pub\_100} &\bf{Cite\_300} &\bf{Pub\_500} &\bf{Pub\_1000}  &Pub\_2000\\
    $\bf{65.7\pm0.6\%\uparrow}$  &  $\bf{67.0\pm0.4\%\uparrow}$  &  $\bf{69.0\pm0.8\%\uparrow}$  & $\bf{72.0\pm0.6\%\uparrow}$ & $68.8\pm0.9\%$ \\
    \bottomrule
  \end{tabular}}
\end{table}
Table 8 reveals that wider is not always better and it depends on the data. We further compare SPIC and state-of-the-art GNN models on well-designed random citation networks in the following. 

\begin{table}
\setlength{\abovecaptionskip}{0pt}
\setlength{\belowcaptionskip}{6pt}
  \caption{Test accuracy (\%) on random citation networks}
  \label{table-9}
  \centering
  \resizebox{\textwidth}{!}{
  \begin{tabular}{llllllll}
    \toprule
    \multicolumn{2}{c}{Model}  &\multicolumn{2}{c}{Cora\_300}&\multicolumn{2}{c}{CiteSeer\_500}&\multicolumn{2}{c}{PubMed\_1000}       \\
    \midrule
   GCN & \bf{DAD(SGC)} & $43.9\pm0.9\%$ & $\bf{74.3\pm0.2\%}$ & $28.7\pm0.6\%$ & $\bf{49.8\pm1.0\%}$ & $42.5\pm1.1\%$ &$\bf{72.0\pm0.6\%}$  \\
   SAGE & \bf{DA} & $50.1\pm0.8\%$ & $\bf{75.3\pm0.6\%}$  & $32.2\pm0.9\%$ & $\bf{50.4\pm1.3\%}$ &$44.2\pm1.5\%$ & $\bf{72.4\pm0.7\%}$ \\
    AGNN & \bf{P\_AGNN} & $42.9\pm0.7\%$ & $\bf{75.0\pm0.5\%}$  & $28.4\pm0.4\%$ & $\bf{49.5\pm0.8\%}$ &$43.5\pm2.4\%$ & $\bf{71.8\pm0.6\%}$ \\
     GAT& \bf{P\_GAT} & $51.9\pm1.6\%$ & $\bf{61.6\pm2.6\%}$  & $33.2\pm0.9\%$ & $\bf{44.2\pm1.3\%}$ &$44.9\pm1.9\%$ & $\bf{70.0\pm 2.8\%}$ \\
    \bottomrule
  \end{tabular}}
\end{table}
Considering Tables 9 and 5, GNNs only well serve the networks with real-world features, whereas SPIC models can still capture communities in these random featured networks. The noteworthy change here is that random features require more iterations ($k=20$).

\subsection{Random Laplacian}
Previous studies focused on the aggregator design and never tested random aggregators. We design  symmetric and asymmetric random aggregator tests on citation networks. In Section 3.1, we define the generalized Laplacian $L^G$ in the manner of message passing. Here, we instantiate it with random symmetric and asymmetric models, 
\begin{equation}
\left\{ 
\begin{array}{l}
RL\_sm \equiv [(H+H^T)/2 + I]^{k}X, \quad H=A*W \\
RL\_am = (A*W + I)^{k}X \\
\end{array}
\right.
,
\end{equation}
where $W$ is a matrix filled with random numbers from a uniform distribution over [0,1).

Another instantiation is to randomly initialize the attention vector of P\_GAT and we propose RGAT\_sm and RGAT\_am.
\begin{table}
\setlength{\abovecaptionskip}{0pt}
\setlength{\belowcaptionskip}{6pt}
  \caption{Test accuracy (\%) of random Laplacian models}
  \label{table-10}
  \centering
  \resizebox{\textwidth}{!}{
  \begin{tabular}{llllllll}
    \toprule
    \multicolumn{2}{c}{Model}  &\multicolumn{2}{c}{Cora}&\multicolumn{2}{c}{CiteSeer}&\multicolumn{2}{c}{PubMed}       \\
    \midrule
   RL\_sm & \bf{RL\_am} & $73.7\pm1.3\%$ & $\bf{75.7\pm1.7\%}$ & $62.7\pm0.9\%$ & $\bf{64.5\pm1.4\%}$ & $75.9\pm0.6\%$ &$\bf{76.9\pm0.3\%}$  \\
   RGAT\_sm & \bf{RGAT\_am} & $80.4\pm0.5\%$ & $\bf{80.2\pm0.7\%}$  & $65.5\pm1.8\%$ & $\bf{65.3\pm2.0\%}$ &$77.2\pm0.7\%$ & $\bf{77.5\pm0.7\%}$ \\
    \bottomrule
  \end{tabular}}
\end{table}

The results in Table 10 are noteworthy, random Laplacian works, which indicates that the topology itself contributes notably to community detection. In this sense, if a method does not outperform the random Laplacian, we may conclude it is not effective enough. The symmetry test on P\_GAT is consistent with the results in Section 4.2, and RL\_am is slightly better than RL\_sm. So, they both work and we need not be concerned about the symmetric or diagonalizable issue of the aggregator. 

\section{Conclusion}
By fully or partly removing activation functions and layer weights of GNNs, we propose subspace power iteration clustering (SPIC) models to explore the mechanism of GNNs. Five cases were discussed: 

 \textcircled{A}All activation functions and layer weights are removed.
 
 \textcircled{B}All layer weights are removed and $ReLU$ is put on the first layer.
 
 \textcircled{C}All layers have activation functions and share the same weight.
 
 \textcircled{D}All layers share the same weight and all activations are removed.
 
 \textcircled{E}Random aggregators are used.

We can extract the power iteration model $M^k X$ in all these cases. Model  \textcircled{A} shows a failure of activations and weights in the case of linear data and random featured graphs. \textcircled{B}, \textcircled{C} and \textcircled{D}suggest that GNNs depend much on activations and layer weights for nonlinear data, but we can follow a power iteration style to remove unnecessary parameters and greatly simplify GNNs. 

The type of data is defined by GAT:
\begin{itemize}
	\item 
	\textbf{Linear:}When running GAT, attention weights almost distribute uniformly regardless of the heads and layers. Activations and layer weights are invalid to some extent on these data.
	\item
	\textbf{Nonlinear:} Significant differences of attention can be observed.
\end{itemize}

It is interesting that random aggregators also work. The topology itself means much to the community detection; perhaps we should not put much focus on the design of GNN aggregators.

Experiments with the above methods verify that GNNs can be simplified to a power iteration style with fewer parameters. The noteworthy improvement is that SPIC models can deal with the random featured networks. Also, the network features can be optimized to speed up GNNs. It is obvious that there is a lot of leeway and creativity in explaining the mechanism of GNN. An interesting direction for future work is to explore the relation between CNN and power iteration.

\section*{Acknowledgments}

The authors would like to thank the anonymous reviewers for their valuable comments and help suggestions that greatly improved the paper’s quality. This work was supported by Natural Science Foundation of China under Grant 61702135, 61571158.

\section*{References}



\begin{thebibliography}{00}


\bibitem{ref1} G. H. Golub and C. F. Van Loan. (2012). ``Symmetric Eigenvalue Problems,'' in {\it Matrix computations}. vol. 3, chap. 8, JHU press.
\bibitem{ref2}F. Lin and W. W. Cohen. (2010). ``Power Iteration Clustering,'' in {\it ICML}, pp. 655-662.
\bibitem{ref3}Y. Gao, X. Yu, and H. Zhang. (2020). ``Uncovering overlapping community structure in static and dynamic networks,'' {\it Knowledge-Based Systems}, pp. 201-202.
\bibitem{ref4}F. Liu, J. Wu, S. Xue, C. Zhou, et al. (2020). ``Detecting the evolving community structure in dynamic social networks,'' {\it World Wide Web}, 23, 715-733.
\bibitem{ref5}F. Liu, J. Wu, C. Zhou, and J. Yang. (2019). ``Evolutionary Community Detection in Dynamic Social Networks,'' {\it IJCNN}, Budapest, Hungary, pp. 1-7.
\bibitem{ref6}F. Liu, S. Xue, J. Wu, C. Zhou, et al. (2020). ``Deep Learning for Community Detection: Progress, Challenges and Opportunities,'' {\it IJCAI}, pp. 4981-4987.
\bibitem{ref7}J. Shi and J. Malik. (2000). ``Normalized cuts and image segmentation,'' {\it PAMI}, 22(8):888–905.
\bibitem{ref8}Q. Li, Z. Han, and X. Wu. (2018). ``Deeper Insights into Graph Convolutional Networks for Semi-Supervised Learning,'' {\it NCAI}, pp. 3538-3545.
\bibitem{ref9}Y. Koren. (2005). ``Drawing graphs by eigenvectors: theory and practice,'' {\it Comput. Math. with Appl.} 49, 1867-1888.
\bibitem{ref10}F. Wu, A. Souza, T. Zhang, C. Fifty, et al. (2019). ``Simplifying Graph Convolutional Networks,'' in {\it ICML}, pp. 6861-6871.
\bibitem{ref11}N. Dehmamy, A. Barabasi, and R. Yu. (2019). ``Understanding the Representation Power of Graph Neural Networks in Learning Graph Topology,'' in {\it NIPS}, pp. 15413-15423.
\bibitem{ref12}R. Ying, D. Bourgeois, J. You, M. Zitnik, and J. Leskovec. (2019). ``GNNExplainer: Generating Explanations for Graph Neural Networks,'' in {\it NIPS}, pp. 9240-9251.
\bibitem{ref13}X. Li and J. Saude. (2020). ``Explain Graph Neural Networks to Understand Weighted Graph Features in Node Classification,'' {\it arXiv:2002.00514 [cs.SI]}.
\bibitem{ref14}Q. Huang, M. Yamada, Y. Tian, D. Singh, et al. (2020). ``GraphLIME: Local Interpretable Model Explanations for Graph Neural Networks,'' {\it arXiv: 2001.06216 [cs.LG]}.
\bibitem{ref15}S. Xie and M. Lu. (2019). ``Interpreting and Understanding Graph Convolutional Neural Network using Gradient-based Attribution Method,'' {\it arXiv: 1903.03768 [cs.LG]}.
\bibitem{ref16}Chung F. (2005). ``Laplacians and the Cheeger Inequality for Directed Graphs,'' {\it Ann. Comb}. 9, 1–19.
\bibitem{ref17}T. N. Kipf and M. Welling. (2017). ``Semi-supervised classification with graph convolutional networks,'' paper presented at {\it ICLR}, Toulon, France.
\bibitem{ref18}W. L. Hamilton, R. Ying, and J. Leskovec. (2017). ``Inductive Representation Learning on Large Graphs,'' paper presented at {\it NIPS}, Long Beach Convention Center, Long Beach.
\bibitem{ref19}J. Bruna J, W. Zaremba, A. Szlam, and Y. LeCun. (2014). ``Spectral Networks and Locally Connected Networks on Graphs,'' paper presented at {\it ICLR}, Banff, Canada.
\bibitem{ref20}M. Defferrard, X. Bresson, and P. Vandergheynst. (2016). ``Convolutional neural networks on graphs with fast localized spectral filtering,'' paper presented at {\it NIPS}, pp. 3844-3852.
\bibitem{ref21}J. Du, S. Zhang, G. Wu, J. M. F. Moura, and S. Kar. (2017). ``Topology adaptive graph convolutional networks,'' {\it arXiv:1710.10370 [cs.LG]}.
\bibitem{ref22}J. Klicpera, A. Bojchevski, and S. Gunnemann. (2019). ``Predict then Propagate: Graph Neural Networks meet Personalized PageRank,'' paper presented at {\it ICLR}, New Orieans, Louisiana.
\bibitem{ref23}K. K. Thekumparampil, S. Oh, C. Wang, and L. J. Li. (2018). ``Attention-based graph neural network for semi-supervised learning,'' {\it arXiv:1803.03735 [stat.ML]}.
\bibitem{ref24}P. Velickovic and G. Cucurull, A. Casanova, A. Romero, et al. (2018). ``Graph Attention Networks,'' paper presented at {\it ICLR}, Vancouver, Canada.
\bibitem{ref25}M. Li, H. Zhang, X. Shi, M. Wang, and Z. Zheng. (2019). ``A Statistical Characterization of Attentions in Graph Neural Networks,'' paper presented at {\it ICLR}, New Orleans, Louisiana.
\bibitem{ref26}P. Sen, G. Namata, M. Bilgic, L. Getoor, et al. (2008). ``Collective Classification in Network Data,'' {\it Ai Mag}. 29, 93-106.
\bibitem{ref27}M. Zitnik and J. Leskovec. (2017). ``Predicting multicellular function through multi-layer tissue networks,'' {\it Bioinformatics}, 33, i190-i198.


\end{thebibliography}


\end{document}